\documentclass[lettersize,journal]{IEEEtran}
\usepackage{amsmath,amsfonts}
\usepackage{algorithmic}
\usepackage{algorithm}
\usepackage{array}
\usepackage[caption=false,font=normalsize,labelfont=sf,textfont=sf]{subfig}
\usepackage{textcomp}
\usepackage{stfloats}
\usepackage{url}
\usepackage{verbatim}
\usepackage{graphicx}
\usepackage{cite}
\usepackage{multicol}
\usepackage{tabularx}
\usepackage{multirow}
\usepackage{booktabs}
\begin{document}

\title{RadioDUN: A Physics-Inspired Deep Unfolding Network for Radio Map Estimation}

\author{Taiqin Chen, Zikun Zhou, Zheng Fang, Wenzhen Zou, Kangjun Liu, Ke Chen, Yongbing Zhang, Yaowei Wang
\thanks{This work was supported in part by the Guangdong Basic and Applied Basic Research Foundation (No. 2025A1515010705).}
  \thanks{Taiqin Chen, Wenzhen Zou, and Yaowei Wang are with the School of Computer Science and Technology, Harbin Institute of Technology (Shenzhen), Shenzhen, SZ 518000 CHN, and also with the Pengcheng Laboratory, Shenzhen, SZ 518000 CHN.}
  \thanks{Yongbing Zhang is with the School of Computer Science and Technology, Harbin Institute of Technology (Shenzhen), Shenzhen, SZ 518000 CHN.}%
  \thanks{Zheng Fang is with the Pengcheng Laboratory, Shenzhen, SZ 518000 CHN, and also with the Department of Computer Science and Engineering, Southern University of Science and Technology, Shenzhen, Shenzhen, SZ 518000 CHN.}
  \thanks{Kangjun Liu, ZiKun Zhou, Ke Chen are with the Pengcheng Laboratory, Shenzhen, SZ 518000 CHN. }
  \thanks{Taiqin Chen and Zikun Zhou contributed equally to this work.}
  \thanks{Corresponding author: Yaowei Wang (wangyaowei@hit.edu.cn)}} 

\markboth{Journal of \LaTeX\ Class Files,~Vol.~14, No.~8, August~2021}%
{Shell \MakeLowercase{\textit{et al.}}: A Sample Article Using IEEEtran.cls for IEEE Journals}


\maketitle

\begin{abstract}
The radio map represents the spatial distribution of spectrum resources within a region, supporting efficient resource allocation and interference mitigation. However, it is difficult to construct a dense radio map as a limited number of samples can be measured in practical scenarios. While existing works have used deep learning to estimate dense radio maps from sparse samples, they are hard to integrate with the physical characteristics of the radio map. To address this challenge, we cast radio map estimation as the sparse signal recovery problem. A physical propagation model is further incorporated to decompose the problem into multiple factor optimization sub-problems, thereby reducing recovery complexity. Inspired by the existing compressive sensing methods, we propose the Radio Deep Unfolding Network (RadioDUN) to unfold the optimization process, achieving adaptive parameter adjusting and prior fitting in a learnable manner. To account for the radio propagation characteristics, we develop a dynamic reweighting module (DRM) to adaptively model the importance of each factor for the radio map. Inspired by the shadowing factor in the physical propagation model, we integrate obstacle-related factors to express the obstacle-induced signal stochastic decay. The shadowing loss is further designed to constrain the factor prediction and act as a supplementary supervised objective, which enhances the performance of RadioDUN. Extensive experiments have been conducted to demonstrate that the proposed method outperforms the state-of-the-art methods. Our code will be made publicly available upon publication.





\end{abstract}

\begin{IEEEkeywords}
Radio map, signal strength prediction, sparse signal recovery, deep unfolding network.
\end{IEEEkeywords}

\section{Introduction}


\begin{figure}[htbp]
\centering
\includegraphics[width=\linewidth]{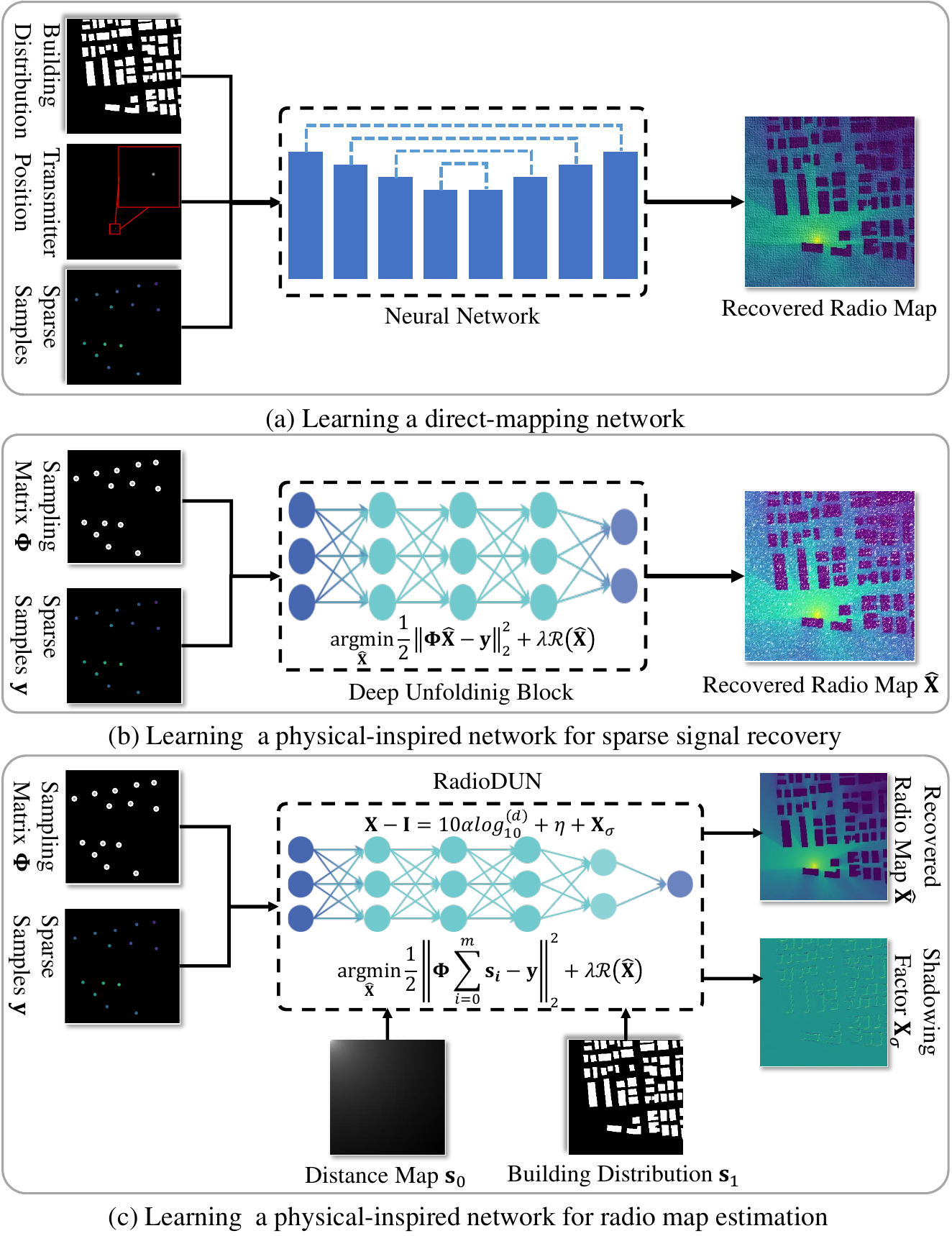}
\caption{Comparison of existing learning-based methods and the proposed method. (a) Existing deep learning-based methods typically utilize a well-designed neural network to construct a direct mapping function of the inputs to the radio map. (b) Existing physical-inspired network for sparse signal recovery employs a deep network to unfold the traditional iterative optimization, which does not account for the influence of environmental information for the radio map. (c) The proposed method integrates the sparse signal recovery with a physical propagation model to derive an iterative solution for radio map estimation and proposes a deep network to unfold the optimization process.}
\label{fig:fig0}
\end{figure}

Radio map~\cite{radio_map} is a visual representation of the spectrum resource that provides the spectrum signal strength at each location within a specific region. The accurate and dense radio maps facilitate spectrum resource allocation and interference management, supporting the reliable operation of wireless applications~\cite{wireless_app}. However, in practical scenarios, there is a limited number of locations where signal strength can be measured. This sparsity in measurement makes the radio map fail to offer effective guiding information for spectrum resource management. To address this issue, Radio Map Estimation (RME)~\cite{RME} has emerged as a crucial research topic, aiming to estimate a dense and accurate radio map based on sparse sampling data.


Conventional RME methods rely on the physical model to generate a radio map by exploiting the effect of physical factors, which can be categorized into the statistical-based and the deterministic-based~\cite{RME_physical}. The former assumes that the signal strength has a log-linear function with physical factors and introduces a Gaussian-distributed shadowing factor to characterize obstacle-induced stochastic signal decay~\cite{RME_sta_1, RME_sta_2}. In contrast, the deterministic-based methods employ ray tracing~\cite{RME_ray_tracing} or finite-difference time-domain simulation~\cite{RME_FDTD} to directly simulate the signal propagation based on the environmental information. Although the above methods achieve efficient or accurate RME based on the physical propagation model, they usually rely on precise transmitter locations, which are difficult to obtain in practice. Moreover, they are unable to balance efficiency and accuracy, which further limits their practical applicability.

Recently, pioneering works~\cite{PMNet, radiounet, rme_gan} introduce deep learning techniques into RME inspired by their remarkable success across various domains~\cite{DL_1, DL_2}. To obtain a finely estimated radio map, environmental factors such as building distribution, terrain map, and so on are used as inputs in addition to the sparse samples~\cite{DS_PP5D}. Subsequently, a learnable neural network is trained on a large-scale dataset to construct the implicit function between inputs and the radio map, which is illustrated in Fig.~\ref{fig:fig0}(a). The network structure typically consists of a sequence of stacked layers like convolution blocks, pooling layers, etc. They are arranged purely to learn a direct mapping from input to output. While these works have achieved notable advancements, the stacked layers solely extract features in the image domain. They neglect the rich physics-informed priors inherent in RME, which constrains their performance~\cite{DUN_1}. Moreover, the construction of an exact mapping from inputs to the radio map demands extensive training data, increasing the difficulty of practical application.


Some works~\cite{RME_CS_1, RME_CS_2} cast RME as the sparse signal recovery and employ compressive sensing (CS)~\cite{CS} theory to derive an iterative optimization for the radio map. Although these methods are enabled to integrate physical characteristics into the recovery procedure, they are unsuitable for practical deployments. In real-world scenarios, the sampling rate is extremely low, which inevitably induces high spatial correlation and violates the inherent requirement of CS~\cite{CS_radio_review}, that the sampling locations exhibit spatial randomness. In digital image recovery, some pioneering works employ the deep unfolding network (DUN) paradigm to unfold the iterative optimization into a trainable deep network, which relaxes the sampling requirement of CS to some extent~\cite{ZYBCS, xie2022puert}. In contrast to digital images, the radio maps are affected by various environmental factors such as obstacle distributions and transmitter locations. Moreover, pixel coordinates of radio maps correspond directly to real-world physical locations, constraining sampling flexibility and further exacerbating the difficulty of RME. However, existing DUN works do not account for the correlation between the radio map and environmental information during optimization, thereby struggling to achieve convincing performance, as shown in Fig.~\ref{fig:fig0}(b).

To address the above challenges, we integrate sparse signal recovery with the physical statistical-based model and derive an iterative optimization for the radio map. Specifically, we cast RME as the sparse signal recovery task, and the spectrum signal is further formulated as the linear combination of multiple factors based on the physical model to reduce the difficulty of recovery. Moreover, to comprehensively model environmental information, we employ the combination of multiple obstacle-related factors to characterize the shadowing factor. Then the alternating optimization (AO)~\cite{alternating_optimization} is employed to decompose the task into multiple optimized subproblems. 

Furthermore, inspired by DUN works, we propose a Radio Deep Unfolding Network (RadioDUN) to replace invariable steps in the optimization such as hyperparameter tuning and prior fitting, illustrated in Fig.~\ref{fig:fig0}(c). The network comprises multiple unfolding blocks, each is strictly equivalent to an iteration in AO. Concretely, an unfolding block consists of three optimized parts, including a gradient descent module (GDM), a dynamic reweighting module (DRM), and a proximal mapping module (PMM). GDM is responsible for updating factors followed by the traditional alternating optimized scheme. Considering the fact that different factors influence the signal differently, DRM is further developed to adaptively model the importance of each factor for the signal and generate an initial radio map. Moreover, the PMM applies a learned proximal operator to the DRM output to enforce the flexible priors, thereby denoising and regularizing the update. Inspired by the conventional solution of the statistical-based model, we design a shadowing loss function to guide RadioDUN in adapting more closely to the signal propagation characteristics in real-world scenarios. Briefly, our contributions are as follows:
\begin{itemize}
    \item We propose a physics-inspired RadioDUN consisting of multiple unfolding blocks to accomplish precise RME, which is designed based on the combination of sparse signal recovery and the physical statistical-based model.
    \item We introduce the DRM to adaptively model the importance of each factor composing the radio map to further align with the characteristics of the spectrum signal.
    \item A shadowing loss function is designed to act as the complementary supervision inspired by the solution of the traditional physical model.
    \item We demonstrate that the proposed RadioDUN achieves state-of-the-art (SOTA) performance through comprehensive experiments.
\end{itemize}

\section{Related Work}
\subsection{Deep learning-based Radio Map Estimation}
Radio map estimation (RME) aims to generate a dense radio map based on input conditions such as obstacle distribution, and sparse measured samples. Owing to the powerful data modeling capabilities of deep learning, numerous deep learning–based methods have been proposed for RME.

For example, Levie et al.~\cite{radiounet} propose a cascaded multistage U-shaped network trained with a large-scale simulation dataset for efficient RME. Zhang et al.~\cite{rme_gan} employ an adversarial learning paradigm, where multiple rounds of generator-discriminator adversarial training are performed to accomplish RME. Lee et al.~\cite{PMNet} introduce multiple dilated convolution blocks~\cite{dilated_conv} with varying dilation rates into a U-shaped network, enlarging the receptive field of the network to further enhance the precision of RME. Despite these methods achieving moderate achievements, their performance is limited due to the incomplete consideration of the physical properties of the radio map. Pioneering work~\cite{phy_RME} has demonstrated that the introduction of a physical model prior in deep learning methods can contribute to performance. They improve network inputs or optimize objective functions solely from a prior-based perspective, making insufficient exploration of the potential of physical models. Therefore, it is necessary to develop a physics-embedded network for RME to comprehensively leverage the physical model.

\subsection{Deep Unfolding Network}
Deep unfolding network (DUN) has become an effective solution for tackling sparse signal recovery problems based on the compressive sensing theory. The core idea of DUN is to unfold the traditional iterative optimization methods into a deep network and to achieve adaptive hyperparameter tuning and prior fitting based on the end-to-end learning paradigm.

Zhang et al.~\cite{CS} unfold the iterative shrinkage-thresholding algorithm (ISTA) into a trainable deep network by mapping the gradient descent and proximal mapping steps of each iteration in ISTA to learnable network layers. Afterwards, many variants emerged for sparse signal recovery problems. For example, Song et al.~\cite{OCTUF} introduce the cross attention mechanism to achieve a lightweight DUN. Li et al.~\cite{DUN_1} further cooperate the priors in the image domain and the convolutional coding domain to tackle the limitations of previous methods with restricted information transmission. Guo et al.~\cite{CPPNet} propose a multi-scale feature fusion module and an iteration fusion strategy to achieve more accurate recovery inspired by the Chambolle and Pock proximal point algorithm. Although these methods have demonstrated outstanding performance in digital image restoration, their direct application to RME without accounting for physical information yields unsatisfactory results.

\section{Method}
In this section, we first introduce sparse signal recovery and the statistical-based physical propagation model, and further derive an alternative iterative optimization for the radio map by integrating them. Then, we present the overall structure of the radio deep unfolding network (RadioDUN). The details of the unfolding block are then introduced. Finally, we present the employed loss function during the training procedure.


\subsection{Theoretical Analysis}
Mathematically, $\mathbf{X}\in \mathbb{R}^{H \cdot W \times 1}$ denotes a radio map that records signal strength at each location, where $H$ and $W$ represent height and width respectively. Then the generation of sparse signal can be formulated as follows:
\begin{equation}
    \mathbf{y} = \mathbf{\Phi} \mathbf{X} + \mathbf{n},
    \label{eq_sam}
\end{equation}
where $\mathbf{\Phi} \in \mathbb{R}^{N \times H \cdot W}$ is a binary sampling matrix, $N$ is the sampling number and $N \ll H \cdot W$,  $\mathbf{y} \in \mathbb{R}^{N \times 1}$ is the sparse signal, $\mathbf{n} \in \mathbb{R}^{N \times 1}$ denotes the observational noise. Correspondingly, the sampling ratio is set as $\frac{N}{H \cdot W}$. 

To accomplish a reconstruction for the original signal, the sparse signal recovery methods usually minimize the objective function:
\begin{equation}
    \arg\min_{\hat{\mathbf{X}}} \frac{1}{2} \|\mathbf{\Phi} \hat{\mathbf{X}} - \mathbf{y}\|_2^2 + \lambda \mathcal{R}(\hat{\mathbf{X}}),
\end{equation}
where $\hat{\mathbf{X}}$ denotes the recovered radio map, $\|\mathbf{\Phi} \hat{\mathbf{X}} - \mathbf{y}\|_2^2$ is the data fidelity term, $\mathcal{R}(\hat{\mathbf{X}})$ is the regularization term that is usually designed based on some specific prior, such as the signal exhibiting sparsity under a specific transform basis~\cite{CS_review}, $\lambda$ is a scaling factor to balance the two terms.


In existing recovery works, the sampling matrix is required to exhibit a high degree of randomness to ensure fine reconstruction quality~\cite{FP_CS}. However, in real-world scenarios, the sampling location is difficult to set flexibly and the sampling ratio may be as low as below $0.01\%$, making the sampling matrix unavoidable to present high spatial correlation. To mitigate the issue, we introduce the physical statistical-based model to express the spectrum signal as the combination of multiple factors. Accordingly, the recovery task is decomposed into multiple sub-tasks, reducing the reconstruction difficulty. Details are as follows.

Taking a classical statistical-based model as an example~\cite{ABG, molisch2012wireless}, spectrum signal can be represented as:
\begin{equation}
    \mathbf{X} - \mathbf{I} = 10 \alpha \mathrm{log}_{10}(d) + \eta + \mathbf{X_\delta},
    \label{equ_phy_model}
\end{equation}
where $\mathbf{I}$ denotes the given signal strength of the transmitter, $d$ denotes the transmitter-receiver (T-R) distance, $\eta$ denotes the fixed loss at reference distance, $\mathbf{X_\delta}$ denotes the Gaussian-distributed shadowing factor that quantifies the random signal fading caused by obstacles, and $\alpha$ is the coefficient. 

In practice, the obstacle-induced stochastic signal decay will vary depending on environmental factors such as obstacle material and distribution, making it difficult to accurately describe using a single variable. Therefore, to enhance the reconstruction precision, we employ a combination of multiple environmental factors $\mathbf{s}_i$ to characterize the shadowing factor, whose number depends on the used dataset, where $i\in[1, \dots, m]$, and $m$ denotes the number of environmental factors. Similarly, we regard the signal decay due to distance as $\mathbf{s}_0$, and the constant term in Eq.~\ref{equ_phy_model} is omitted for simplicity. Therefore, we reformulate Eq.~\ref{equ_phy_model} as a more general form, which is expressed as:
\begin{equation}
    \mathbf{X} = \sum_{i=0}^{m} \mathbf{s}_{i}.
    \label{equ_sim_phy}
\end{equation}

Then the objective function can be rewritten based on Eq.~\ref{equ_sim_phy}, which is demonstrated as:
\begin{equation}
    \arg\min_{\hat{\mathbf{X}}} \frac{1}{2} \|\mathbf{\Phi} \sum_{i=0}^{m} \mathbf{s}_{i} - \mathbf{y}\|_2^2 + \lambda \mathcal{R}(\hat{\mathbf{X}}),
\end{equation}

Furthermore, the alternating optimization~\cite{alternating_optimization} is adapted to recover the original signal, which decomposes the optimized process into $m+1$ sub-parts. The solving procedure is formulated as: 
\begin{align}
    \mathbf{s}_i^{k} &= \mathbf{s}_i^{k-1} 
        - \beta_i \mathbf{\Phi}^\top ( 
            \mathbf{\Phi} (\, 
                \sum_{j=0}^{i-1} \mathbf{s}_j^{k} 
                + \sum_{l=i}^{m} \mathbf{s}_l^{k-1} 
            \,) 
            - \mathbf{y} 
        ),\label{equ_gd} \\ 
    \mathbf{X}^{k} &= \sum_{i=0}^m \mathbf{s}_i^{k} + \lambda \mathcal{R}(\mathbf{\hat{X}}), \label{equ_solve}
\end{align}
where $\mathbf{s}_i^{k}$ denotes the $i$-th environmental factor at the $k$-th iteration, $\beta_i$ is the scaling coefficient, $i \in [0, \dots,m]$, $\mathbf{\Phi}^\top$ denotes the transpose of $\mathbf{\Phi}$.

In traditional optimized works, it is common to design regularization terms $\mathcal{R}(\mathbf{\hat{X}})$ based on handcrafted priors to improve recovery performance. For example, natural images are usually sparse in transform domains such as wavelet and gradient domains. However, predefined priors fail to accommodate complex and changing scenarios in RME. Additionally, traditional optimization requires careful parameter selection and suffers a heavy computational burden~\cite{MST}. To tackle these challenges, we propose a radio deep unfolding network (RadioDUN) to unfold the iterative optimization shown in Eq.~\ref{equ_solve}, achieving a more generalized and robust term fitting and adaptive parameter selection by the deep network with an end-to-end learning manner.

\renewcommand{\dblfloatpagefraction}{.9}
\begin{figure*}[htbp]
\centering
\includegraphics[width=\linewidth]{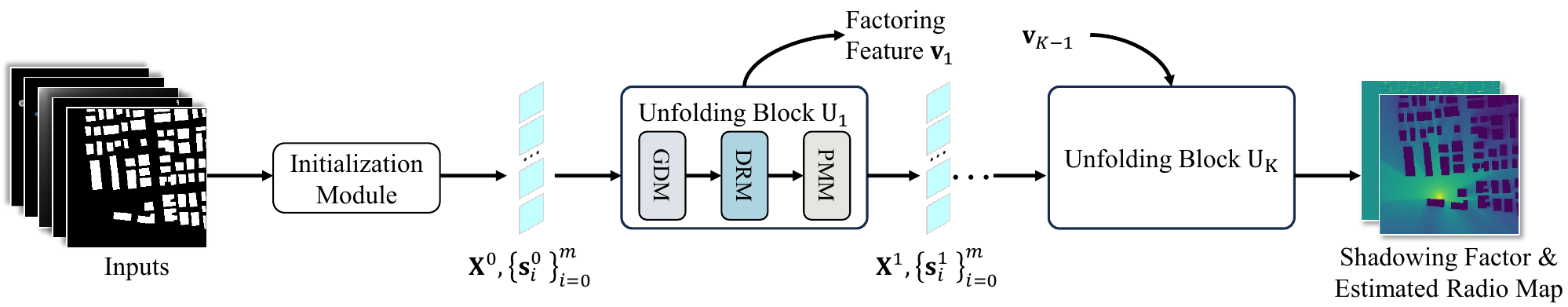}
\caption{Overview of the proposed radio deep unfolding network (RadioDUN), which unfolds the optimization shown in Eq.~\ref{equ_solve} by $K$ unfolding blocks. The sparse radio map and the environmental information such as building distribution are fed into RadioDUN along with the sampling matrix $\mathbf{\Phi}$. An initialization module is employed to generate an initial value for Eq.~\ref{equ_solve}, denoting as $\mathbf{X}^{0}, \mathbf{s}_{0}^{0}, \dots, \mathbf{s}_{m}^{0}$, where $m$ represents the number of factors, depending on the applied dataset. Then $K$ unfolding blocks including gradient descent module (GDM), dynamic reweighting module (DRM), and proximal mapping module (PMM) are utilized to update each component, which the $i$-th unfolding block is denoted as $\mathrm{U}_i$. For $\mathrm{U}_i$, except for the updated factors $\{\mathbf{s}_j^{i}\}_{j=0}^m$ and the radio map $\mathbf{X}^i$, a factoring feature $\mathbf{v}_i$ is also yielded to support $\mathrm{U}_{i+1}$ in adaptively modeling the importance of each factor for $\mathbf{X}^{i+1}$. Through $K$ cascade blocks, the predicted radio map $\mathbf{\hat{X}}$ and is acquired. In addition, we integrate obstacle-related environmental factors to generate the shadowing factor $\mathbf{X}_\delta$ for the complementary supervision.}
\label{fig:fig1}
\end{figure*}

\subsection{Overall Structure of RadioDUN}
The proposed radio deep unfolding network (RadioDUN) is illustrated in Fig.~\ref{fig:fig1}. Except for sparse samples $\mathbf{y}$ and the sampling matrix $\mathbf{\Phi}$, RadioDUN also takes the distribution of transmitters or obstacles as input to describe the environmental information in the specific areas, whose values are binary to indicate presence or absence. The binary form is hard to serve as environmental factors in Eq.~\ref{equ_sim_phy} to compose the radio map. Therefore, an initialization module is employed to acquire initial values for each factor. Then $K$ unfolding blocks that strictly correspond to the traditional optimization are further cascaded to obtain updated factors and the predicted radio map. Moreover, we generate the shadowing factor to execute additional supervision by integrating obstacle-related environmental factors.

\subsection{Initialization Module}
First, a distance map is generated according to the transmitter position, which is formulated as:
\begin{equation}
    \mathbf{P}_0(x_r,y_r) = 1-\mathrm{Norm}(\mathrm{log}_{10}^{1+\sqrt{(x_t-x_r)^2+(y_t-y_r)^2}}),
    \label{equ_dis}
\end{equation}
where $\mathbf{P}_0 \in \mathbb{R}^{1 \times H \times W}$ denotes the distance map, $x_r$ and  $y_r$ represent the coordinates of pixels, and $x_r \in [1,H]$, $y_r \in [1,W]$, $x_t$ and  $y_t$ represent coordinates of the transmitter, $\mathrm{Norm}(\cdot)$ denotes the min-max normalized function for normalizing values to $[0,1]$.

Similarly, we denote the distribution of obstacles as $\mathbf{P}_i, i\in [1, \dots, m]$, where $m$ represents the number of inputs, depending on the used dataset. Further, we employ two consecutive convolution blocks to obtain the initial values for each factor. The conversion is expressed as:
\begin{equation}
    \mathbf{s}_i^0 = \mathrm{Conv}_2(\mathrm{Cat}(\mathbf{y}, \mathbf{s}_0^0, \dots, \mathbf{s}_{i-1}^0, \mathbf{P}_i, \dots, \mathbf{P}_m)),
\end{equation}
where $\mathbf{s}_i^0 \in \mathbb{R}^{1 \times H \times W}$ denotes the initial value of $i$-th factor that makes up the radio map, $i \in [0, m]$, $\mathrm{Conv}_2(\cdot)$ denotes two consecutive convolution blocks with batch normalization and activation function, $\mathrm{Cat}(\cdot)$ denotes the function that concatenates inputs in the channel dimension, $\mathbf{P}_i \in \mathbb{R}^{1 \times H \times W}$ denotes the $i$-th input.

Then we integrate all initial factors through a convolution block whose kernel size is $1\times1$ to obtain the initial radio map $\mathbf{X}^0$, which is formulated as:
\begin{equation}
    \mathbf{X}^0 = \mathrm{Conv}_{1\times1}(\mathrm{Cat}(\{\mathbf{s}_i^0\}_{i=0}^m),
\end{equation}
where $\mathrm{Conv}_{1\times1}(\cdot)$ denotes the convolution block with $1\times1$ kernel size.


\subsection{Unfolding Block}
The unfolding block strictly corresponds to one iteration of the alternating optimization, which is composed of the gradient descent module (GDM), the dynamic reweighting module (DRM), and the proximal mapping module (PMM), whose structures are illustrated in Fig.~\ref{fig:fig2}. We take the $k$-th unfolding block as an example to introduce the details.

\subsubsection{Gradient Descent Module}
\begin{figure*}[htbp]
\centering
\includegraphics[width=\linewidth]{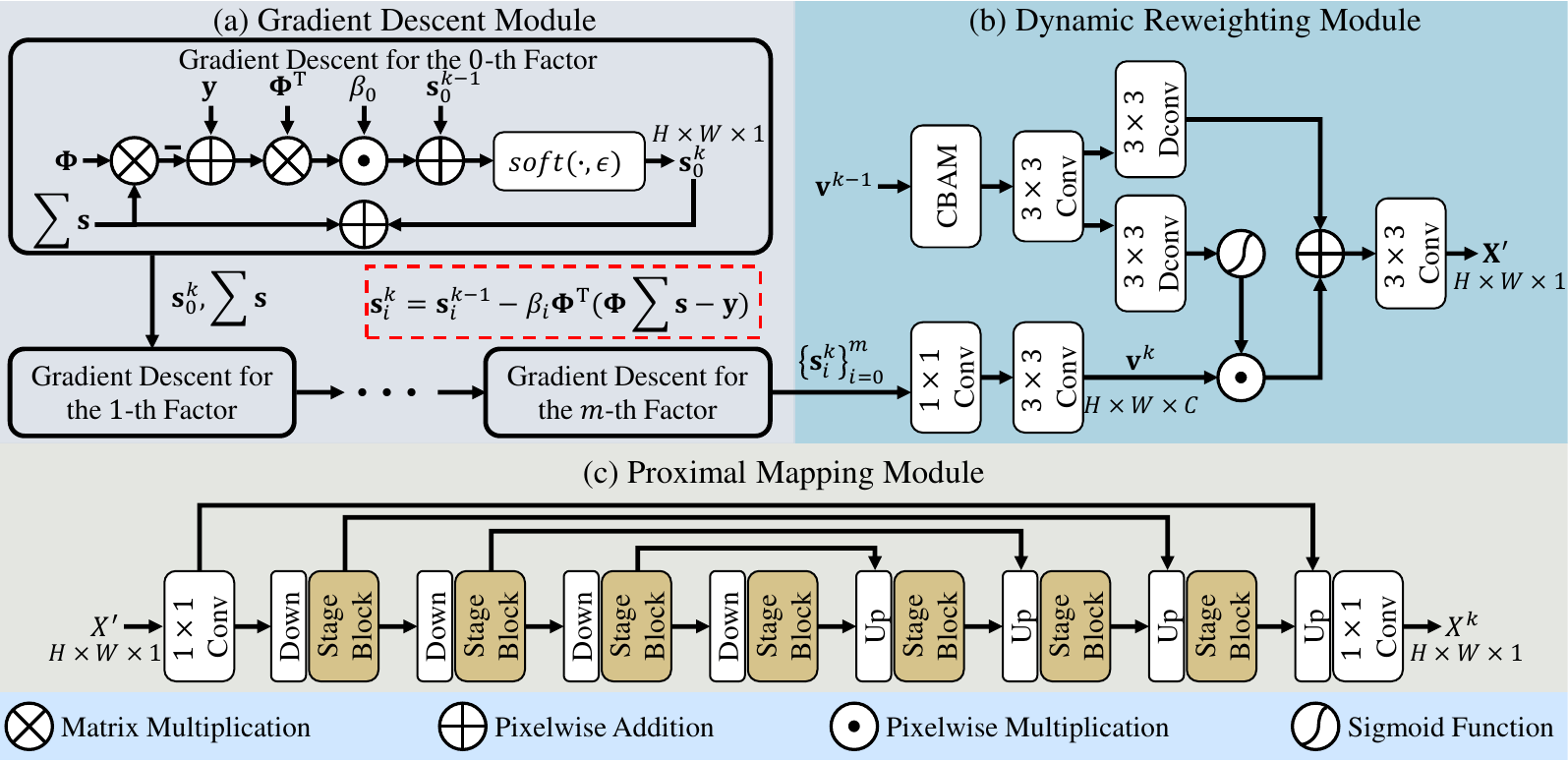}
\caption{Details of the unfolding block, which consists of a gradient descent module (GDM), a dynamic reweighting module (DRM), and a proximal mapping module (PMM). The $k$-th unfolding block is taken as the example. (a) An illustration of GDM, which is responsible for updating each factor alternately following Eq.~\ref{equ_gd}. (b) An illustration of DRM, which dynamically models the importance of the update factors and integrates them to obtain the rough radio map $\mathbf{X}^{'}$. (c) An illustration of PMM, which is employed to fit generalized priors for radio map estimation and outputs a fine recovered radio map.}
\label{fig:fig2}
\end{figure*}

To be free from cumbersome hyperparameter tuning, we develop the GDM, whose structure is depicted in Fig.~\ref{fig:fig2}(a). GDM employs learnable parameters to replace the hyperparameters in Eq.~\ref{equ_gd}. Further, we follow the classical iterative shrinkage-thresholding algorithm~\cite{ISTA} to add a soft thresholding operation after each factor update to achieve sparsification of the solution while ensuring global convergence. Therefore, the update for each factor can be formulated as:
\begin{align}
    \mathbf{s}_i^k &= \mathrm{soft}(\mathbf{s}_i^{k-1} - \beta_i \mathbf{\Phi}^\top (\mathbf{\Phi} \sum \mathbf{s} - \mathbf{y}), \epsilon), \\
    \mathrm{soft}(\mathbf{z}, \epsilon) &= \mathrm{sign}(\mathbf{z}) \cdot \mathrm{max}(|\mathbf{z}|-\epsilon, 0),\\
    \sum \mathbf{s} &= \sum_{j=0}^{i-1}\mathbf{s}_j^{k} + \sum_{l=i}^m \mathbf{s}_l^{k-1},
\end{align}
where $\mathbf{s}_i^k$ denotes the $i$-th factor in the $k$-th iteration, $\mathrm{soft}(\cdot, \epsilon)$ denotes the soft thresholding operation, parameters $\epsilon, \beta_i$ are learnable, $\mathrm{sign}(\cdot)$ is the function that specifies the sign of the variable, and its return values are $[1, 0, -1]$, corresponding to positive, zero, and negative inputs, respectively.


\renewcommand{\dblfloatpagefraction}{.9}

\subsubsection{Dynamic Reweighting Module}
The update for each factor is completed through GDM. However, their importance for the radio map is not comprehensively explored, which is against the fact that different factors have distinct effects on the radio map. To tackle this issue, we develop a dynamic reweighting module (DRM) to adaptively model the importance of each factor for the reconstructed map and output the rough radio map, whose structure is illustrated in Fig.~\ref{fig:fig2}(b).

Since single-channel images have a limited ability to express information, we concatenate all factors along the channel dimension, and further transform them into a unified factoring feature $\mathbf{v}^{k} \in \mathbb{R}^{C \times H \times W}$ through a group of convolution blocks. Subsequently, we employ the previous factoring feature $\mathbf{v}^{k-1}$ as guidance to extract differential distributional features, thereby constructing implicit importance relationships between factors. 

Concretely, a convolutional block attention module (CBAM)~\cite{CBAM} first generates channel-wise and spatial attention maps, which are then used to reweight $\mathbf{v}^{k-1}$. CBAM performs channel realignment and spatial refocusing on $\mathbf{v}^{k-1}$ respectively, exploring the representative information in the features effectively. $\mathbf{v}^{k-1}$ is first squeezed through the spatial average pooling layer and the spatial maximum pooling layer. Then a learnable linear layer is employed to generate the channel-wise importance map, which is expressed as follows:
\begin{equation}
    \mathbf{C}_\mathrm{ch} = \sigma(\mathrm{Linear}(\mathrm{Ave}_\mathrm{S}(\mathbf{v}^{k-1})+\mathrm{Max}_\mathrm{S}(\mathbf{v}^{k-1})), \\
\end{equation}
where $\mathbf{C}_\mathrm{ch} \in \mathbb{R}^{C \times 1 \times 1}$ denotes the channel-wise importance map, $\sigma(\cdot)$ denotes the sigmoid function, $\mathrm{Linear}(\cdot)$ denotes the learnable linear layer, $\mathrm{Ave}_\mathrm{S}(\cdot)$ and $\mathrm{Max}_\mathrm{S}(\cdot)$ denote the average and maximum pooling layers along the spatial dimension.

Then, $\mathbf{v}^{k-1}$ is reweighted by the channel-wise importance map $\mathbf{C}_\mathrm{ch}$ and the resulting feature $\mathbf{v}_\mathrm{ch}^{k-1}$ is modulated by the spatial importance map $\mathbf{C}_\mathrm{sp}$ to emphasize important areas. The generation of $\mathbf{C}_\mathrm{sp}$ is formulated as follows:
\begin{equation}
    \mathbf{C}_\mathrm{sp} = \sigma(\mathrm{Conv}_\mathrm{7\times7}(\mathrm{Cat}(\mathrm{Ave}_\mathrm{C}(\mathbf{v}_\mathrm{ch}^{k-1}), \mathrm{Max}_\mathrm{C}(\mathbf{v}_\mathrm{ch}^{k-1})))), 
\end{equation}
where $\mathrm{Conv}_\mathrm{7\times7}(\cdot)$ represents the convolution block with $7\times7$ kernel size, $\mathrm{Ave}_\mathrm{C}(\cdot)$ and $\mathrm{Max}_\mathrm{C}(\cdot)$ represent the average and maximum pooling layers along the channel dimension.

A convolution block with $3\times3$ kernel size is then used to refine the representative information of the modulated feature $\mathbf{v}^{'}$. Furthermore, an adaptive information selection is accomplished by convolution blocks to integrate $\mathbf{v}^{'}$ and $\mathbf{v}^{k}$, thereby producing the rough radio map $X^{'}$. The procedure is expressed as:
\begin{equation}
    \mathbf{X}^{'} = \mathrm{Conv}_{3\times3}(\mathbf{v}^{k}\cdot \sigma(\mathrm{DConv}(\mathbf{v}^{'}))+\mathrm{DConv}(\mathbf{v}^{'})),
\end{equation}
where $\mathrm{Conv}_{3\times3}(\cdot)$ denotes the convolution block with $3\times3$ kernel size, $\mathrm{DConv}(\cdot)$ denotes the depthwise convolution block with $3\times3$ kernel size~\cite{dwconv}.

\subsubsection{Proximal Mapping Module}

For Eq.~\ref{equ_solve}, a regularization term based on the priors for the radio map is designed to improve the reconstruction performance. In the traditional optimized methods, it is common to employ predefined priors such as a sparse constraint in the defined transform domain. However, it may excessively smooth fine details and demand extensive manual tuning of parameters~\cite{CS_denoiser}. To tackle this issue, we develop a proximal mapping module (PMM) to adaptively learn flexible and robust regularization terms for RME. The structure of PMM is depicted in Fig.~\ref{fig:fig2}(c).

PMM takes the rough radio map $\mathbf{X}^{'}$ as the input, and adopts a symmetric U-shaped deep network~\cite{unet} based on an encoder-decoder structure to explore the refined radio map $\hat{\mathbf{X}}^{k}$. The encoder progressively downsamples the input to extract multiscale contextual features, and the decoder subsequently upsamples these features to restore spatial resolution and reconstruct fine details. During the restoring procedure, skip connections are inserted between the encoder and the decoder to transfer corresponding features from the encoder to the decoder to integrate high-resolution and semantic information, thereby enhancing reconstruction accuracy.

\begin{figure}[htbp]
\centering
\includegraphics[width=\linewidth]{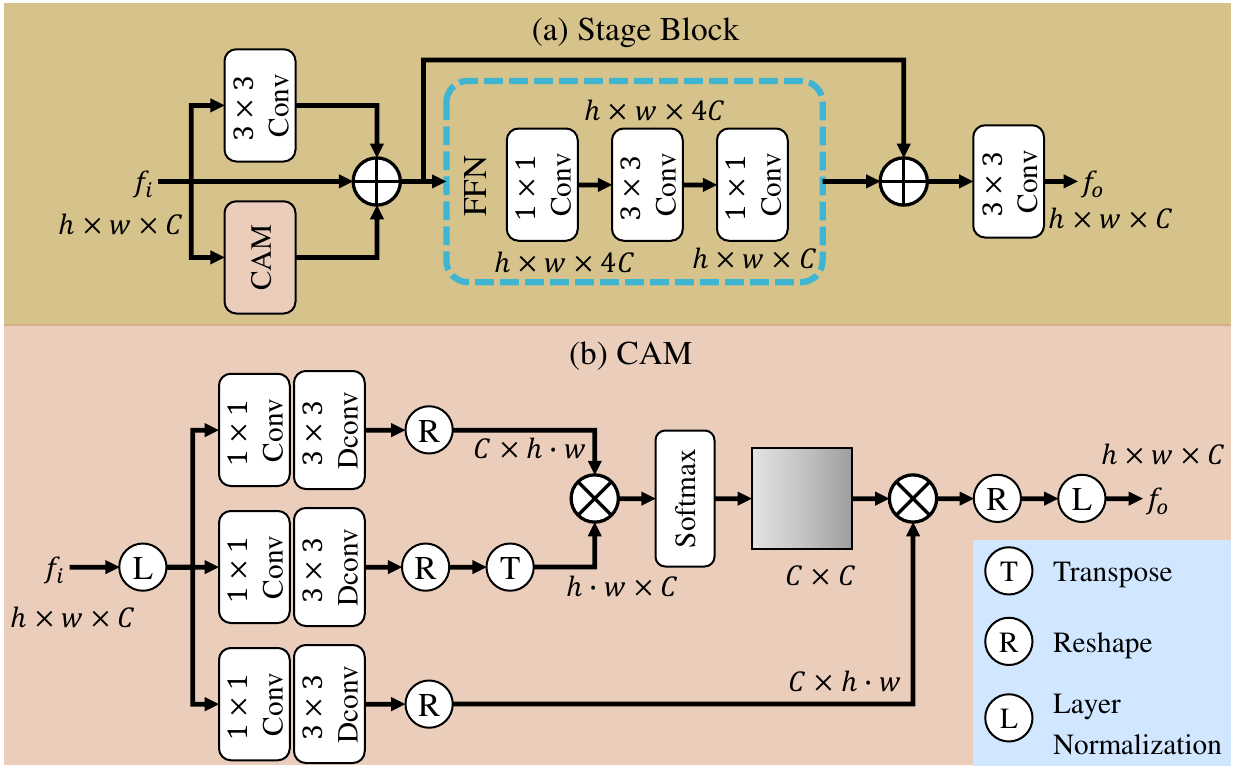}
\caption{Details of the stage block, which is the major component of PMM. (a) The overall architecture of the stage block. (b) The structure of channel-wise attention module (CAM).}
\label{fig:fig3}
\end{figure}

Further, to construct the long-range dependencies within features while preserving effective perception of local regions, we develop the stage block as major components of PMM. The stage block adopts a hybrid convolution and attention structure~\cite{cao2024hybrid}, whose structure is illustrated in Fig.~\ref{fig:fig3}(a). 

The input feature is processed by a convolution block and a channel-wise attention module (CAM) in parallel, where the former extracts local semantic information along the spatial dimension and the latter explores long-range dependencies along the channel dimension. Then, the extracted features are fused through pixelwise addition, and the semantic differences in the fused feature are aligned using a feed-forward network (FFN) consisting of three convolution blocks. To avoid gradient vanishing and accelerate model convergence, we add shortcut connections in the stage block. 

The detailed structure of CAM is shown in Fig.~\ref{fig:fig3}(b). CAM first normalizes the input feature $f_\mathrm{i}$ by a layer normalization, followed by the generation of three tokens: query $\mathbf{Q} \in \mathbb{R}^{C \times h \cdot w}$, key $\mathbf{K} \in \mathbb{R}^{C \times h \cdot w}$, and value $\mathbf{V} \in \mathbb{R}^{C \times h \cdot w}$ in the same manner, where $h$ and $w$ denote the spatial size of the feature. Taking query $\mathbf{Q}$ as an example, the generating procedure is expressed as:
\begin{equation}
\mathbf{Q} = \mathrm{Re}(\mathrm{DConv}(\mathrm{Conv}_{1\times1}(\mathrm{LN}(f_\mathrm{i})))), 
\end{equation}
where $\mathrm{Re}(\cdot)$ denotes the reshape operator, $\mathrm{LN}(\cdot)$ denotes the layer normalization.

After that, an attention map $\mathbf{A} \in \mathbb{R}^{C \times C}$ depicting the importance distribution of the feature in the channel dimension is computed based on the matrix multiplication between the query token and the transposed key token. The calculation of attention map is formulated as follows:
\begin{equation}
     \mathbf{A} = \mathrm{Softmax}(\mathbf{Q}\mathbf{K}^\top), 
\end{equation}
where $\mathrm{Softmax}(\cdot)$ represents the function that maps input into the range $(0,1)$, $\mathbf{K}^\top$ represents the transpose of key $\mathbf{K}$. At last, value $\mathbf{V}$ is reweighted by the attention map, and outputs the refocused feature $f_\mathrm{o}$.

\subsection{Loss Function}
Inspired by the solution of the statistical-based model~\cite{ABG}, the shadowing loss is proposed to serve as the additional supervised function. Specifically, the variables in Eq.~\ref{equ_phy_model} can be simplified as $\mathbf{E} = \mathbf{X} - \mathbf{I}$ and $\mathbf{F}=10\mathrm{log}_{10}(d)$. Correspondingly, the shadowing factor $\mathbf{X}_\delta$ is formulated as:
\begin{equation}
    \mathbf{X}_\delta = \mathbf{E} - \alpha \mathbf{F} - \eta.
\end{equation}
Since $\mathbf{X}_\delta$ is modeled as a parameter conforming to the zero-mean Gaussian distribution, its standard deviation $\sigma_{\mathbf{X}_\delta}$ can be expressed as:
\begin{equation}
    \sigma_{\mathbf{X}_\delta} = \sqrt{\frac{\sum(\mathbf{E}-\alpha \mathbf{F} - \eta)^2}{N}}.
    \label{equ_std}
\end{equation}

Eq.~\ref{equ_std} also characterizes the deviation of the actual observations from the model predictions. Hence, minimizing $\sigma_{\mathbf{X}_\delta}$ is equivalent to minimizing the fitting error of the statistical-based model. Based on the theory, we introduce a shadowing loss $\mathcal{L}_{\sigma}$, which is formulated as:
\begin{align}
    \mathcal{L}_{\sigma} &= \frac{(\hat{\mathbf{X}}_\sigma - \overline{\hat{\mathbf{X}}_\sigma})^2}{H \times W} + \frac{(\mathbf{X}^{\mathrm{GT}} - \hat{\mathbf{X}} + \hat{\mathbf{X}}_\sigma - \overline{\hat{\mathbf{X}}_\sigma})^2}{H \times W},\\
    \hat{\mathbf{X}}_\sigma &= \mathrm{Conv}_{1\times1}(\mathrm{Cat}(\mathbf{s}_1,\dots,\mathbf{s}_m)), \label{equ_shadowing}
\end{align}
where $\hat{\mathbf{X}}_\sigma$ denotes the shadowing factor, acquired by integrating obstacle-related environmental factors, $\overline{\hat{\mathbf{X}}_\sigma}$ denotes the mean of $\hat{\mathbf{X}}_\sigma$, $\mathbf{X}^{\mathrm{GT}}$ denotes the ideal radio map, $\hat{\mathbf{X}}$ denotes the recovered radio map through RadioDUN.

Similar to other works, the mean square error $\mathcal{L}_{\mathrm{MSE}}$ is taken as the fidelity term to supervise the learning procedure of RadioDUN. $\mathcal{L}_{\mathrm{MSE}}$ is expressed as:
\begin{equation}
    \mathcal{L}_{\mathrm{MSE}} = \frac{(\mathbf{X}^{\mathrm{GT}} - \hat{\mathbf{X}})^2}{H \times W}.
\end{equation}
Hence, the total loss function $\mathcal{L}_\mathrm{T}$ is formulated as:
\begin{equation}
    \mathcal{L}_\mathrm{T} = \mathcal{L}_{\sigma} + \mu \mathcal{L}_{\mathrm{MSE}},
\end{equation}
where $\mu$ is a scaling parameter, which is set to $1$ by default.

\section{Experiment}

In this section, extensive experiments are conducted to validate the proposed method. We first present the experimental settings, including the employed datasets, evaluation metrics, and implementation details. To demonstrate the advanced performance of the proposed method, we conduct a comparative experiment with existing radio map estimation (RME) methods and sparse signal recovery methods. Further, an experiment with a limited amount of training data is conducted to quantify the sensitivity of each method to changes in training data size. Subsequently, a varying number of samples study is conducted to indicate the sensitivity of each method in terms of the number of samples. Moreover, a transferability study is performed to demonstrate the potential of our method in cross-scenarios. Finally, an ablation study is performed to comprehensively demonstrate the effectiveness of the proposed modules.
\subsection{Experimental Setup}
\subsubsection{Dataset}
We employ the widely adopted dataset RadioMapSeer~\cite{radiounet} to evaluate the performance of each method. RadioMapSeer consists of $700$ city maps, whose spatial resolution is $256\times256$. Further, the simulated radio maps are generated through the Dominant Path Model (DPM)~\cite{DPM} method and Intelligent Ray Tracing with $4$ interactions (IRT4)~\cite{IRT}, respectively. For DPM, each city map is simulated with $80$ different transmitter locations, while merely $2$ different locations are considered for IRT4 due to the heavy computational resources. Except for the transmitter location, the environmental inputs in the dataset contain the distribution of buildings and cars, which are the binary maps to indicate the obstacle locations.

\subsubsection{Evaluation Metrics}
To quantitatively evaluate method performance, root mean square error (RMSE), structural similarity index measure (SSIM), and peak signal-to-noise ratio (PSNR) are utilized as evaluation metrics. Specifically, RMSE is used to measure the difference between the recovered radio map of each method and the ideal radio map, whose calculation is formulated as:
\begin{equation}
    \mathrm{RMSE} = \sqrt{\frac{1}{n}\sum_{i=1}^{n}(\mathbf{X}^\mathrm{GT}_i - \hat{\mathbf{X}}_i)^2},
\end{equation}
where $n$ denotes the number of testing samples, $\mathbf{X}^\mathrm{GT}_i$ denotes the ground truth of $i$-th sample, and $\hat{\mathbf{X}}_i$ denotes the recovered map of $i$-th sample.

SSIM is a metric used to assess image quality, which is commonly employed to measure the similarity between reconstructed and ideal images in tasks such as image compression, denoising, etc~\cite{SSIM}. Likewise, we utilize the metric to quantify the similarity between the recovered and ideal radio map. Taking the $i$-th sample to introduce the computing procedure, which is expressed as:
\begin{equation}
\mathrm{SSIM} = \frac{(2\mu_\mathrm{g}\mu_\mathrm{r} + C_1)(2\sigma_{\mathrm{gr}} + C_2)}
{(\mu_\mathrm{g}^2 + \mu_\mathrm{r}^2 + C_1)(\sigma_\mathrm{g}^2 + \sigma_\mathrm{r}^2 + C_2)},
\end{equation}
where $\mu_\mathrm{g}$ and $\sigma_{\mathrm{g}}$ represent the average value and the standard deviation for the ground truth respectively, $\mu_\mathrm{r}$ and $\sigma_{\mathrm{r}}$ represent the similar meanings for the recovered map, $\sigma_{\mathrm{gr}}$ represents the covariance between the recovered and ideal radio map, $C_1$ and $C_2$ are constant terms to prevent instability due to a denominator with a tiny value.

We also employ PSNR to evaluate the quality of recovery by calculating the pixel-level error between the ideal map and the prediction, which is formulated as:
\begin{equation}
    \mathrm{PSNR} = 10 \times \mathrm{log}_{10}(\frac{C_\mathrm{M}^2}{\frac{1}{n}\sum_{i=1}^{n}(\mathbf{X}^\mathrm{GT}_i - \hat{\mathbf{X}}_i)^2}), 
\end{equation}
where $C_\mathrm{M}$ denotes the maximum value of the radio map.

\subsubsection{Implementation Details}
All experiments are conducted on the Pytorch platform, whose version is $1.12.0$ and the computing hardware is an NVIDIA RTX 4090D GPU. The chosen optimizer is AdamW with the weight decay $1\times 10^{-4}$, and the initial learning rate is $1\times 10^{-3}$. In the training procedure, we employ the cosine annealing strategy to dynamically adjust the learning rate. The number of training epochs is set as $100$ for each method. In addition, we normalize the strength of radio maps for all data to range $[0,1]$.

\begin{table*}[htbp]
\centering
\caption{Quantitative results on two RME datasets. The optimal results are denoted as bold and the sub-optimal results are denoted as underlined.}
\begin{tabular}{*{13}{c}}

  \toprule
  \multirow{3}*{Methods} & \multicolumn{6}{c}{Transmitter-unknown condition} & \multicolumn{6}{c}{Transmitter-known condition} \\
  \cmidrule(lr){2-7}\cmidrule(lr){8-13} &
  \multicolumn{3}{c}{DPM} & \multicolumn{3}{c}{IRT4} & \multicolumn{3}{c}{DPM} & \multicolumn{3}{c}{IRT4} \\  
  \cmidrule(lr){2-4}\cmidrule(lr){5-7}\cmidrule(lr){8-10}\cmidrule(lr){11-13}
  & RMSE$\downarrow$ & SSIM$\uparrow$ & PSNR$\uparrow$ & RMSE$\downarrow$ & SSIM$\uparrow$ & PSNR$\uparrow$ & RMSE$\downarrow$ & SSIM$\uparrow$ & PSNR$\uparrow$ & RMSE$\downarrow$ & SSIM$\uparrow$ & PSNR$\uparrow$ \\
  \midrule
        RadioUNet & 0.0363 & 0.9463 & \underline{30.5446} 
        & \underline{0.0647} & \underline{0.8620} & \underline{24.2210}
        & \underline{0.0096} & \textbf{0.9803} & 40.3138
        & \underline{0.0362} & \underline{0.9067} & \underline{28.8311}\\
        PMNet & \underline{0.0320} & \textbf{0.9498} & 30.0752
        & 0.1382 & 0.6992 & 17.2343
        & 0.0098 & 0.9796 & \underline{40.3509}
        & 0.1610 & 0.6934 & 15.8693 \\
        RME-GAN & 0.0728 & 0.5528 & 23.0345
        & 0.2150 & 0.2798 & 13.3525
        & 0.0308 & 0.8509 & 30.2352  
        & 0.2196 & 0.2854 & 13.1845\\
        OCTUF & 0.1075 & 0.8020 & 19.3818
        & 0.1117 & 0.7955 & 19.0498
        & 0.0829 & 0.8138 & 21.6325
        & 0.0976 & 0.7793 & 20.2193\\
        Ours & \textbf{0.0298} & \underline{0.9478} & \textbf{31.4449}
        & \textbf{0.0387} & \textbf{0.8915} & \textbf{28.2542}
        & \textbf{0.0094} & \underline{0.9798} & \textbf{40.5116}
        & \textbf{0.0190} & \textbf{0.9407} & \textbf{34.4558}\\
  \bottomrule
\end{tabular}

\label{tab_compar}
\end{table*}

\subsubsection{Compared Methods}
To demonstrate the advantage of our method, multiple state-of-the-art (SOTA) methods are taken as comparisons, which can be classified into deep learning-based RME methods and sparse signal recovery methods. The former consists of RadioUNet~\cite{radiounet}, PMNet~\cite{PMNet}, and RME-GAN~\cite{rme_gan}. The latter includes OCTUF~\cite{OCTUF}. The details of these methods are as follows:
\begin{itemize}
    \item RadioUNet consists of a cascade of two multi-stage U-shaped networks, each consisting of a repeated stack of convolution blocks and downsampling layers.
    \item PMNet adopts the encoder-decoder architecture and integrates dilated convolution blocks~\cite{dilated_conv} with varying dilation rates to enlarge the receptive field.
    \item RME-GAN is built upon an adversarial learning paradigm, leveraging the competitive interplay between a generator and a discriminator to accomplish RME.
    \item OCTUF employs a deep unfolding network (DUN) paradigm to perform digital image restoration under sparse sampling, whose unfolding blocks are designed based on a cross-attention mechanism.
\end{itemize}

\subsection{Comparison Study}
To ensure fairness, all compared methods are implemented strictly following the descriptions in their papers or using their open-source code. The number of training epochs is configured as $100$ for all methods. For the proposed method and OCTUF, the number of unfolding blocks $K$ is configured as $3$.

In the comparison study, DPM and IRT4 are employed as the evaluated datasets. Each dataset is divided into training, validation, and testing sets with a ratio $0.75:0.05:0.2$. Considering the difficulty of measuring sample acquisition in real scenarios, we limit the number of samples to $9$ in our experiments. To account for practical situations where the transmitter location is agnostic, we further assess the performance of each method under transmitter-unknown conditions. In this case, we assume the transmitter is at $(0,0)$ to generate the distance map. 

Results are demonstrated in Tab.~\ref{tab_compar}. Compared with the comparisons, the proposed method achieves the optimal results in terms of RMSE, SSIM, and PSNR, indicating the significant advantages of our method in RME. Specifically, for deep learning-based RME methods, the availability of transmitter position substantially enhances the performance in the case of adequate training data. For OCTUF, the performance gain is much smaller than other compared methods since it does not account for the effect of environmental information on the radio map. However, deep learning-based RME methods struggle to effectively construct a direct mapping from the inputs to the radio map. This weakness is more serious for PMNet and RME-GAN due to the larger number of parameters or the adversarial training strategy. In contrast to the comparisons, we comprehensively take into account the environmental information during the iterative optimization, which reformulates the radio map as a combination of multiple environmental factors based on a physical propagation model. Secondly, the dynamic reweighting module adaptively adjusts the importance of each factor to the radio map, further aligning with the physical properties of signal propagation. Thirdly, the shadowing loss provides an additional optimized objective from the signal fading perspective, enhancing the performance of our method. Facilitating the above-mentioned designs, our method obtains the best results. Under the transmitter-unknown condition, the proposed method achieves a $6.87\%$ improvement in RMSE on DPM and a $40.19\%$ improvement on IRT4 compared to the sub-optimal method.

\begin{table*}[htbp]
\centering
\caption{Quantitative results when the training data size is varied on DPM.}
\begin{tabular}{*{13}{c}}

  \toprule
  \multirow{2}*{Methods} & \multicolumn{4}{c}{$2800$} & \multicolumn{4}{c}{$5600$} & \multicolumn{4}{c}{$8400$} \\  
  \cmidrule(lr){2-5}\cmidrule(lr){6-9}\cmidrule(lr){10-13}
  & RMSE$\downarrow$ & SSIM$\uparrow$ & PSNR$\uparrow$ & Dec(\%)$\downarrow$ & RMSE$\downarrow$ & SSIM$\uparrow$ &  PSNR$\uparrow$ & Dec(\%)$\downarrow$ & RMSE$\downarrow$ & SSIM$\uparrow$ & PSNR$\uparrow$ & Dec(\%)$\downarrow$  \\
  \midrule
        RadioUNet & \underline{0.0579} & \underline{0.9127} & \underline{24.7755} & {59.50}
        & 0.0496 & 0.9257 & 26.1316 & 36.64
        & 0.0459 & 0.9319 & 26.7937 & 26.45\\
        PMNet & 0.0965 & 0.8722 & 20.3326 & 201.56
        & \underline{0.0431} & \underline{0.9281} & \underline{27.3910} & {34.69}
        & \underline{0.0400} & \underline{0.9336} & \underline{28.0481} & \underline{25.00} \\
        RME-GAN & 0.2271 & 0.2808 & 13.2436 & 211.95
         & 0.2264 & 0.2841 & 13.2742 & 210.99
         & 0.2267 & 0.2845 & 13.2677 & 211.40\\
        OCTUF & 0.1129 & 0.8218 & 18.9526 & \textbf{5.02}
        & 0.1120 & 0.8236 & 19.0213 & \textbf{4.19}
        & {0.1113} & 0.8229 & {19.0772} & \textbf{3.53}\\
        Ours & \textbf{0.0451} & \textbf{0.9209} & \textbf{27.0169} & \underline{51.34}
        & \textbf{0.0396} & \textbf{0.9322} & \textbf{28.1613} & \underline{32.89}
        & \textbf{0.0373} & \textbf{0.9361} & \textbf{28.6934} & {25.17} \\
  \bottomrule
\end{tabular}

\label{tab2}
\end{table*}

\renewcommand{\dblfloatpagefraction}{.9}
\begin{figure*}[htbp]
\centering
\includegraphics[width=\linewidth]{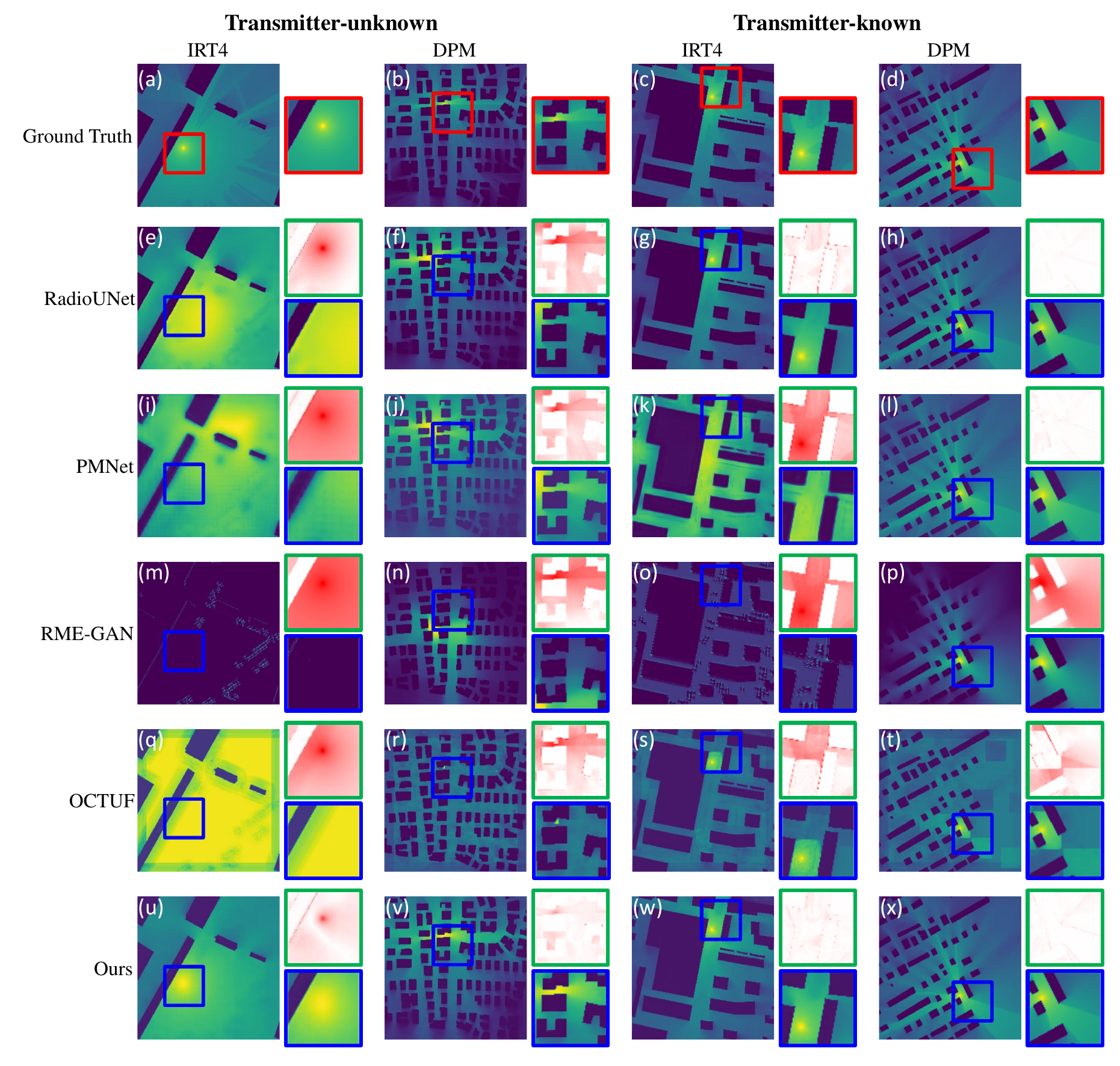}
\caption{The visual comparison with other SOTA methods. A shift toward yellow denotes higher signal strength. Regions around the strength maximum are enlarged, indicated by blue rectangular boxes. Further, the corresponding error distribution maps are also shown, indicated by green rectangular boxes. A shift toward red denotes higher error value.}
\label{fig:fig4}
\end{figure*}

We also visualize one scene in each experimental setting, which is shown in Fig.~\ref{fig:fig4}. To enable a more intuitive comparison between methods, we zoom in on the regions surrounding the strength maximum and present corresponding error distribution maps, which are marked by the blue rectangle and the green rectangle. By comparing with the ground truth, it is found that the positions of the strength maximum predicted by existing methods exhibit a significant shift when the transmitter position is missing, as shown in the enlarged areas of Fig.~\ref{fig:fig4}(e), (f), (i), and (j). In contrast, the proposed method mitigates this issue by incorporating physical priors, as shown in the enlarged areas of Fig.~\ref{fig:fig4}(u) and (v). When the transmitter position is known but the training data is insufficient, the prediction of existing methods will be too smooth to reflect the signal fading caused by obstacles, causing a higher error value compared to the proposed method, as shown in the third column in Fig.~\ref{fig:fig4}. In a nutshell, the proposed method yields predictions that are closer to the ground truth compared to the comparative methods.

\subsection{Training Data Size Sensitivity Study}
Further, we conduct a training data size sensitivity study to demonstrate the performance variation for each method. In the experiment, the transmitter location is unknowable, and the number of samples is set to $9$. DPM is taken as the experimental dataset, and the number of training data is configured as $2800$, $5600$, and $8400$, respectively. In addition to the evaluated metrics, we also measure the performance degradation compared to the complete training data condition under RMSE for each method, which is expressed as:
\begin{equation}
    \mathrm{Dec} = \frac{\mathrm{RMSE}_{\mathrm{all}}-\mathrm{RMSE}_{\mathrm{reduced}}}{\mathrm{RMSE}_{\mathrm{all}}} \times 100\%,
\end{equation}
where $\mathrm{RMSE}_{\mathrm{all}}$ denotes the RMSE under the complete training data, and $\mathrm{RMSE}_{\mathrm{reduced}}$ denotes the RMSE under the reduced training data.

Results are indicated in Tab.~\ref{tab2}. Compared to the comparative methods, the proposed method achieves the best results under the limited training data size condition. Specifically, PMNet exhibits a strong dependency on the training data size, with its performance degrading significantly when the amount of data is insufficient. This issue is more serious for RME-GAN since the employed adversarial learning paradigm requires a larger training dataset to ensure network convergence. Although OCTUF does not exhibit significant degradation in accuracy when the training data is limited, its performance is still untrustworthy. In contrast to existing methods, we incorporate the physical propagation model to reformulate RME as a joint optimization over multiple factors, which decreases the predicted difficulty. Moreover, we introduce the physical properties of the radio map into the proposed method, facilitating the model convergence under challenging conditions. Leveraging these physics-informed insights, our method achieves robust prediction accuracy with limited data and exhibits less performance degradation than the comparisons when training data is reduced. As seen in Tab.~\ref{tab2}, compared with RadioUNet and PMNet in terms of Dec, our method is enhanced by $8.16\%$ on the $2800$ training data over the sub-optimal method, and a boost of $1.80\%$ on the $5600$ training data.




\subsection{Varying Number of Samples Study}

We further evaluate the performance of each method across varying numbers of samples. The experimental dataset is configured as the size-constrained DPM with $8400$ training data, and the transmitted position is unknown during the study. The number of samples is set as $\{t^2\}_{t=3}^{10}$, respectively.

\begin{table}[htbp]
\centering
\caption{Quantitative results when the number of samples is varied.}

\scriptsize
\begin{tabular}{ccccccc}
\toprule
\multirow{2}*{Methods} & \multicolumn{3}{c}{\textbf{9}} & \multicolumn{3}{c}{\textbf{16}} \\
\cmidrule(lr){2-4} \cmidrule(lr){5-7} 
 & RMSE$\downarrow$ & SSIM$\uparrow$ & PSNR$\uparrow$ 
 & RMSE$\downarrow$ & SSIM$\uparrow$ & PSNR$\uparrow$ \\
\midrule
RadioUNet & 0.0459 & 0.9319 & 26.7937 & 0.0348 & 0.9418 & 29.2171  \\
PMNet & \underline{0.0400} & \underline{0.9336} & \underline{28.0481} & \underline{0.0295} & \underline{0.9436} & \underline{30.6605} \\
RME-GAN & 0.2267 & 0.2845 & 13.2677 & 0.2267 & 0.2851 & 13.2348  \\
OCTUF & 0.1113 & 0.8229 & 19.0772 & 0.1018 & 0.7985 & 19.8523  \\
Ours & \textbf{0.0373} & \textbf{0.9361} & \textbf{28.6934} & \textbf{0.0285} & \textbf{0.9456} & \textbf{31.6101} \\
\bottomrule

\multirow{2}*{Methods} & \multicolumn{3}{c}{\textbf{25}} & \multicolumn{3}{c}{\textbf{36}} \\
\cmidrule(lr){2-4} \cmidrule(lr){5-7} 
 & RMSE$\downarrow$ & SSIM$\uparrow$ & PSNR$\uparrow$ 
 & RMSE$\downarrow$ & SSIM$\uparrow$ & PSNR$\uparrow$ \\
\midrule
RadioUNet & 0.0293 & 0.9475 & 30.7079 & 0.0262 & 0.9508 & 31.6654  \\
PMNet & \textbf{0.0232} & \textbf{0.9530} & \underline{32.7374} & \underline{0.0215} & \underline{0.9554} & \underline{33.3921} \\
RME-GAN & 0.2266 & 0.2861 & 13.2378 & 0.2267 & 0.2854 & 13.2363  \\
OCTUF & 0.0910 & 0.8070 & 20.8317 & 0.0837 & 0.8157 & 21.5593  \\
Ours & \underline{0.0235} & \underline{0.9528} & \textbf{33.0843} & \textbf{0.0202} & \textbf{0.9572} & \textbf{34.3274} \\
\bottomrule

\multirow{2}*{Methods} & \multicolumn{3}{c}{\textbf{49}} & \multicolumn{3}{c}{\textbf{64}} \\
\cmidrule(lr){2-4} \cmidrule(lr){5-7} 
 & RMSE$\downarrow$ & SSIM$\uparrow$ & PSNR$\uparrow$ 
 & RMSE$\downarrow$ & SSIM$\uparrow$ & PSNR$\uparrow$ \\
\midrule
RadioUNet & 0.0243 & 0.9524 & 32.3200 & 0.0231 & 0.9538 & 32.7421  \\
PMNet & \underline{0.0194} & \underline{0.9575} & \underline{34.2493} & \underline{0.0174} & \underline{0.9612} & \underline{35.2172} \\
RME-GAN & 0.2266 & 0.2850 & 13.2371 & 0.2265 & 0.2869 & 13.2417  \\
OCTUF & 0.0769 & 0.8226 & 22.2874 & 0.1102 & 0.8063 & 19.1678  \\
Ours & \textbf{0.0184} & \textbf{0.9606} & \textbf{35.1088} & \textbf{0.0167} & \textbf{0.9630} & \textbf{35.8857} \\
\bottomrule

\multirow{2}*{Methods} & \multicolumn{3}{c}{\textbf{81}} & \multicolumn{3}{c}{\textbf{100}} \\
\cmidrule(lr){2-4} \cmidrule(lr){5-7} 
 & RMSE$\downarrow$ & SSIM$\uparrow$ & PSNR$\uparrow$ 
 & RMSE$\downarrow$ & SSIM$\uparrow$ & PSNR$\uparrow$ \\
\midrule
RadioUNet & 0.0219 & 0.9558 & 33.2044 & 0.0211 & 0.9570 & 33.5125  \\
PMNet & \underline{0.0194} & \underline{0.9548} & \underline{34.2769} & \underline{0.0191} & \underline{0.9578} & \underline{34.3998} \\
RME-GAN & 0.2266 & 0.8482 & 23.9151 & 0.0584 & 0.8596 & 24.6918  \\
OCTUF & 0.0638 & 0.9122 & 25.4649 & 0.0343 & 0.9261 & 29.1651  \\
Ours & \textbf{0.0155} & \textbf{0.9661} & \textbf{36.5473} & \textbf{0.0150} & \textbf{0.9661} & \textbf{36.7876} \\
\bottomrule
\end{tabular}
\label{tab_var}
\end{table}

Results are indicated in Tab.~\ref{tab_var}. From signal recovery insights, increasing the number of samples usually facilitates the quality of predicted radio maps. However, PMNet achieves optimal performance when the number of samples is set to 64, whereas OCTUF exhibits a significant performance degradation under the same condition. Further, the increase in the number of samples does not free RME-GAN from its inability to converge due to insufficient training data, whose accuracy remains essentially unchanged. In contrast to the comparisons, the proposed method achieves the optimal results, and the performance is enhanced as the number of samples increases.



\subsection{Transferability Study}
In real-world scenarios, large-scale datasets are often unavailable, and training a model from scratch demands excessive time and computational resources. To tackle this challenge, transferring a pre-trained model to the target scenario has become a popular approach~\cite{TL}. Therefore, we conduct a transferability study to demonstrate the generalization performance of each method in the cross-scenario settings.

Following the above experiment, we configure the transmitter position as unknown, and the number of samples is $9$. Each method is trained in the DPM and then transferred into the IRT4 to evaluate its performance, where DPM has $40$ times more data than IRT4. We first verify the performance when the pre-trained model is applied directly to IRT4, denoted as zero-shot validation. Except for the zero-shot setting, part of the training data of IRT4 is employed to update the parameters of the pre-trained model, whose ratio is configured as $10\%$ and $30\%$, respectively. The number of training epochs is fixed at $20$ to simulate the fast iterations in practical applications.

\begin{figure}[htbp]
\centering
\includegraphics[width=\linewidth]{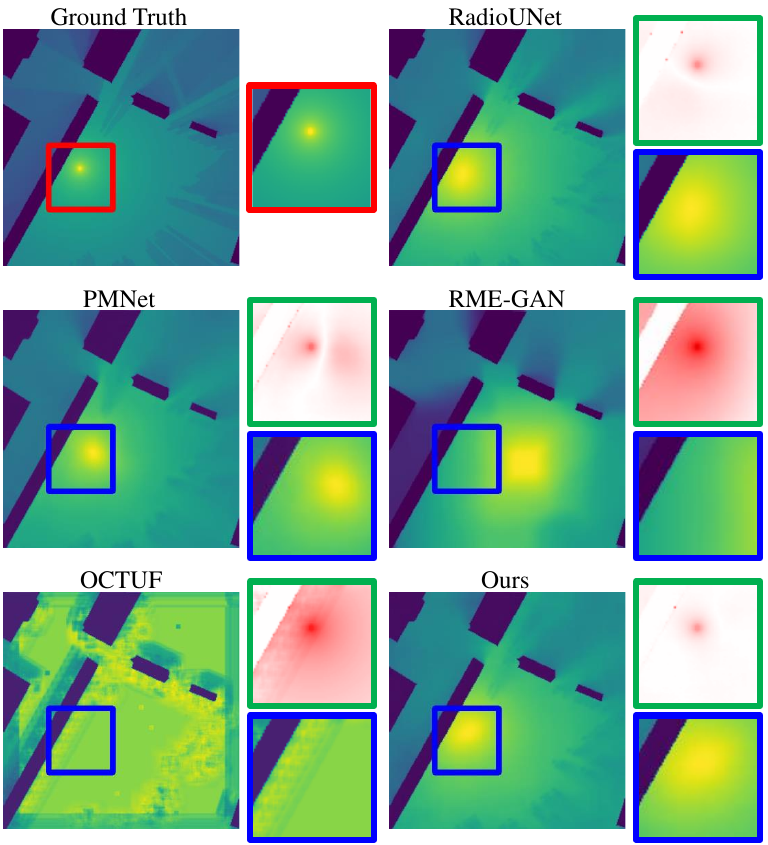}
\caption{Illustration of prediction for each method in the cross-scenario setting. Each method is pre-trained on the DPM with 100 epochs, then fine-tuned on the IRT4 with $30\%$ data, while setting the number of training epochs to $20$.}
\label{fig:fig5}
\end{figure}

\begin{figure}[htbp]
\centering
\includegraphics[width=\linewidth]{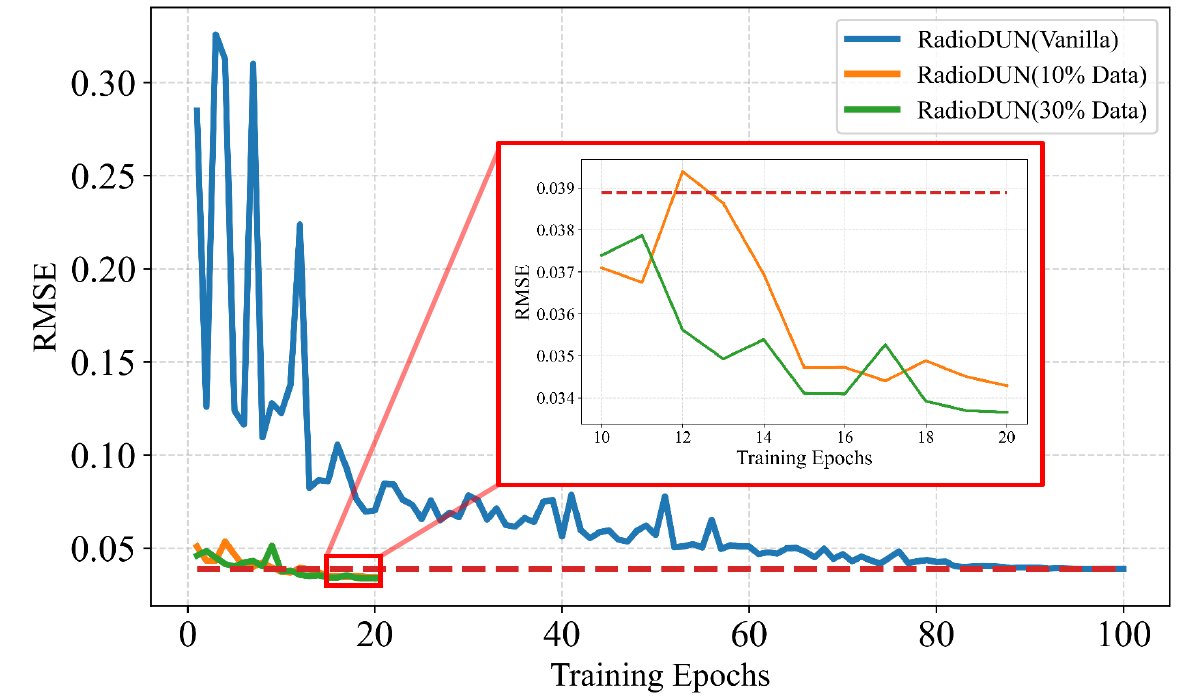}
\caption{Comparison of the training efficiency of the proposed method on the IRT4 with and without the utilization of the pre-trained model.}
\label{fig:fig6}
\end{figure}

Results are indicated in Tab.~\ref{tab_zero_shot}. The proposed method achieves optimal performance in the above experimental settings. In the zero-shot condition, our method achieves a $13.50\%$ improvement in RMSE compared to the sub-optimal method. As the training data of the target scenario increases, the proposed method converges to better accuracy. In the $30\%$ training data, the proposed method is boosted by $16.94\%$ in terms of RMSE over the sub-optimal method. Furthermore, we visualize predictions of each method under the $30\%$ training data setting, which is illustrated in Fig.~\ref{fig:fig5}. The regions around the strength maximum are enlarged and the corresponding error distribution maps are presented, as shown in the blue rectangle and green rectangle in Fig.~\ref{fig:fig5} respectively. It is observed from the error map comparison, the prediction of our method is closer to the ground truth compared with the competing methods. Moreover, we depict a comparison of the training efficiency of the proposed method with and without the involvement of a pre-trained model, as shown in Fig.~\ref{fig:fig6}. Results indicate that our method can converge significantly faster with the assistance of the pre-trained model, achieving superior performance compared to the vanilla results. These results demonstrate that the proposed method provides better generalization capabilities compared to the existing methods while maintaining excellent prediction performance in the cross-scenario setting. 




\begin{table*}[htbp]
\centering
\caption{Quantitative results on the cross-scenario settings. Each method is first trained on the DPM, then transferred into the IRT4.}
\begin{tabular}{*{10}{c}}

  \toprule
  \multirow{2}*{Methods} & \multicolumn{3}{c}{Zero-shot} & \multicolumn{3}{c}{10\%} & \multicolumn{3}{c}{30\%}  \\  
  \cmidrule(lr){2-4}\cmidrule(lr){5-7}\cmidrule(lr){8-10}
  & RMSE$\downarrow$ & SSIM$\uparrow$ & PSNR$\uparrow$ & RMSE$\downarrow$ & SSIM$\uparrow$ & PSNR$\uparrow$& RMSE$\downarrow$ & SSIM$\uparrow$ & PSNR$\uparrow$\\
  \midrule
        RadioUNet & 0.0589 & 0.8731 & 25.7258 
        & {0.0551} & {0.8830} & 26.5347
        & 0.0496 & 0.8844 & 26.9645 \\
        PMNet & \underline{0.0563} & \underline{0.8782} & \underline{25.0419}
        & \underline{0.0446} & \underline{0.8987} & \underline{27.9970}
        & \underline{0.0425} & \underline{0.9010} & \underline{28.4856}\\
        RME-GAN & 0.0742 & 0.5421 & 22.6231
        & 0.0736 & 0.5372 & 22.7210
        & 0.0771 & 0.5390 & 22.3325\\
        OCTUF & 0.1112 & 0.7621 & 19.0999
        & 0.1101 & 0.7048 & 19.7772
        & 0.0954 & 0.7458 & 20.6746 \\
        Ours & \textbf{0.0487} & \textbf{0.8826} & \textbf{26.3073}
        & \textbf{0.0412} & \textbf{0.9008} & \textbf{28.5371}
        & \textbf{0.0353} & \textbf{0.9133} & \textbf{29.5011}
        \\
  \bottomrule
\end{tabular}

\label{tab_zero_shot}
\end{table*}

\subsection{Ablation Study}
We conduct an ablation study to verify the effectiveness of the proposed modules, such as the dynamic reweighting module (DRM) and the shadowing loss (SL). Following the above experimental settings, the transmitter position is hidden and the number of samples is kept as $9$. The employed dataset is the size-constrained DPM with $5600$ training data. In the experiment, we utilize a convolution block with $1\times1$ kernel size as the counterpart for DRM, and SL is eliminated to demonstrate its influence on our method. 

Results of ablation study are shown in Tab.~\ref{tab_abla}. DRM can effectively enhance the performance of RME when integrated individually. However, the prediction accuracy remains essentially unchanged when leveraging SL alone. This may be attributed to the absence of modeling the importance of environmental factors for the radio map. As a result, the shadowing factor generated by Eq.~\ref{equ_shadowing} is biased, providing limited supervised information. Therefore, the performance is further improved by $4.12\%$ compared to the vanilla method when combining DRM and SL. These results comprehensively identify the effectiveness of the proposed modules.

\begin{table}[ht]
    \centering
    \caption{Ablation study of Dynamic Reweighting Module (DRM) and Shadowing Loss (SL).}
    \begin{tabular}{cc ccc}
        \toprule
        DRM & SL 
        & RMSE$\downarrow$ & SSIM$\uparrow$ & PSNR$\uparrow$ \\
        \midrule
        $\times$ & $\times$ & 0.0413 & 0.9271 & 27.7875 \\
        $\times$ & $\checkmark$ & 0.0414 & 0.9291 & 27.7782 \\
        \checkmark & $\times$ & \underline{0.0403} & \underline{0.9303} & \underline{28.0170} \\
        \checkmark & \checkmark & \textbf{0.0396} & \textbf{0.9322} & \textbf{28.1613} \\
        \bottomrule
    \end{tabular}
    
    \label{tab_abla}
\end{table}

Additionally, we investigate the relationship between the number of unfolding blocks $K$ and model performance. Except for metrics to quantify accuracy, we also measure the number of parameters for model (Para) and the giga floating-point operations per second (GFLOPs). Results are presented in Tab.~\ref{tab_abla_NS}. Contrary to expectations, the model performance does not consistently improve with an increasing $K$. This may be attributed to the fact that a larger $K$ introduces more parameters, demanding more training data to optimize comprehensively. Therefore, the choice of $K$ should take into account the number of parameters and the size of the training data. We identify the optimal number of unfolding blocks as $3$ in our experimental setup.


\begin{table}[ht]
    \centering
    \caption{Ablation study for the number of the unfolding blocks $K$.}
    \begin{tabular}{c ccccc}
        \toprule
        $K$ 
        & RMSE$\downarrow$ & SSIM$\uparrow$ & PSNR$\uparrow$ & Para(Mb) $\downarrow$ & GFLOPs $\downarrow$ \\
        \midrule
        2 & 0.0404 & \underline{0.9303} & \underline{28.8860} & \textbf{9.62} & \textbf{26.84} \\
        3 & \textbf{0.0396} & \textbf{0.9322} & 28.1613 & \underline{14.43} & \underline{40.44} \\
        4 & \underline{0.0398} & {0.9300} & \textbf{28.9427} & 19.24 & 54.04 \\
        5 & {0.0411} & {0.9279} & {28.6618} & 24.05 & 67.64 \\
        \bottomrule
    \end{tabular}
    
    \label{tab_abla_NS}
\end{table}

\section{Conclusion}
In this paper, we propose a physics-inspired deep unfolding network for radio map estimation (RME), denoted as RadioDUN. Specifically, we cast RME as the sparse signal recovery problem and formulate the solution as an iterative optimization. Furthermore, the statistics-based physical propagation model is introduced to decompose the optimization into multiple factor recovery sub-problems to reduce the difficulty of RME. Inspired by the compressive sensing methods, we propose RadioDUN to unfold the above-mentioned optimization, avoiding the cumbersome hyperparameter tuning and overwhelming prior fitting. To leverage the signal propagation characteristics, we develop the dynamic reweighting module to adaptively model the importance of each factor for the radio map. Moreover, the shadowing loss is designed to guide the optimization for the proposed method. Comprehensive experiments on two RME datasets are conducted, demonstrating that the proposed method outperforms state-of-the-art methods in diverse experimental settings.


\bibliographystyle{IEEEtran}
\bibliography{Bibliography}

\begin{thebibliography}{10}
\providecommand{\url}[1]{#1}
\csname url@samestyle\endcsname
\providecommand{\newblock}{\relax}
\providecommand{\bibinfo}[2]{#2}
\providecommand{\BIBentrySTDinterwordspacing}{\spaceskip=0pt\relax}
\providecommand{\BIBentryALTinterwordstretchfactor}{4}
\providecommand{\BIBentryALTinterwordspacing}{\spaceskip=\fontdimen2\font plus
\BIBentryALTinterwordstretchfactor\fontdimen3\font minus \fontdimen4\font\relax}
\providecommand{\BIBforeignlanguage}[2]{{%
\expandafter\ifx\csname l@#1\endcsname\relax
\typeout{** WARNING: IEEEtran.bst: No hyphenation pattern has been}%
\typeout{** loaded for the language `#1'. Using the pattern for}%
\typeout{** the default language instead.}%
\else
\language=\csname l@#1\endcsname
\fi
#2}}
\providecommand{\BIBdecl}{\relax}
\BIBdecl

\bibitem{radio_map}
D.~Romero and S.-J. Kim, ``Radio map estimation: A data-driven approach to spectrum cartography,'' \emph{IEEE Signal Processing Magazine}, vol.~39, no.~6, pp. 53--72, 2022.

\bibitem{wireless_app}
M.~Z. Chowdhury, M.~Shahjalal, S.~Ahmed, and Y.~M. Jang, ``6g wireless communication systems: Applications, requirements, technologies, challenges, and research directions,'' \emph{IEEE Open Journal of the Communications Society}, vol.~1, pp. 957--975, 2020.

\bibitem{RME}
D.~Romero, T.~N. Ha, R.~Shrestha, and M.~Franceschetti, ``Theoretical analysis of the radio map estimation problem,'' \emph{IEEE Transactions on Wireless Communications}, 2024.

\bibitem{RME_physical}
{\c{C}}.~Yapar, F.~Jaensch, R.~Levie, G.~Kutyniok, and G.~Caire, ``Overview of the first pathloss radio map prediction challenge,'' \emph{IEEE Open Journal of Signal Processing}, 2024.

\bibitem{RME_sta_1}
C.~Phillips, D.~Sicker, and D.~Grunwald, ``A survey of wireless path loss prediction and coverage mapping methods,'' \emph{IEEE Communications Surveys \& Tutorials}, vol.~15, no.~1, pp. 255--270, 2012.

\bibitem{RME_sta_2}
S.~Sun, T.~S. Rappaport, T.~A. Thomas, A.~Ghosh, H.~C. Nguyen, I.~Z. Kovacs, I.~Rodriguez, O.~Koymen, and A.~Partyka, ``Investigation of prediction accuracy, sensitivity, and parameter stability of large-scale propagation path loss models for 5g wireless communications,'' \emph{IEEE transactions on vehicular technology}, vol.~65, no.~5, pp. 2843--2860, 2016.

\bibitem{RME_ray_tracing}
Z.~Yun and M.~F. Iskander, ``Ray tracing for radio propagation modeling: Principles and applications,'' \emph{IEEE access}, vol.~3, pp. 1089--1100, 2015.

\bibitem{RME_FDTD}
R.~A. Marshall, T.~Wallace, and M.~Turbe, ``Finite-difference modeling of very-low-frequency propagation in the earth-ionosphere waveguide,'' \emph{IEEE Transactions on Antennas and Propagation}, vol.~65, no.~12, pp. 7185--7197, 2017.

\bibitem{PMNet}
J.-H. Lee and A.~F. Molisch, ``A scalable and generalizable pathloss map prediction,'' \emph{IEEE Transactions on Wireless Communications}, 2024.

\bibitem{radiounet}
R.~Levie, {\c{C}}.~Yapar, G.~Kutyniok, and G.~Caire, ``Radiounet: Fast radio map estimation with convolutional neural networks,'' \emph{IEEE Transactions on Wireless Communications}, vol.~20, no.~6, pp. 4001--4015, 2021.

\bibitem{rme_gan}
S.~Zhang, A.~Wijesinghe, and Z.~Ding, ``Rme-gan: A learning framework for radio map estimation based on conditional generative adversarial network,'' \emph{IEEE Internet of Things Journal}, vol.~10, no.~20, pp. 18\,016--18\,027, 2023.

\bibitem{DL_1}
Y.~Guo, Y.~Liu, A.~Oerlemans, S.~Lao, S.~Wu, and M.~S. Lew, ``Deep learning for visual understanding: A review,'' \emph{Neurocomputing}, vol. 187, pp. 27--48, 2016.

\bibitem{DL_2}
D.~W. Otter, J.~R. Medina, and J.~K. Kalita, ``A survey of the usages of deep learning for natural language processing,'' \emph{IEEE transactions on neural networks and learning systems}, vol.~32, no.~2, pp. 604--624, 2020.

\bibitem{DS_PP5D}
S.~Zhang, S.~Jiang, W.~Lin, Z.~Fang, K.~Liu, H.~Zhang, and K.~Chen, ``Generative ai on spectrumnet: An open benchmark of multiband 3d radio maps,'' \emph{IEEE Transactions on Cognitive Communications and Networking}, 2024.

\bibitem{DUN_1}
W.~Li, B.~Chen, S.~Liu, S.~Zhao, B.~Du, Y.~Zhang, and J.~Zhang, ``D 3 c 2-net: Dual-domain deep convolutional coding network for compressive sensing,'' \emph{IEEE Transactions on Circuits and Systems for Video Technology}, 2024.

\bibitem{RME_CS_1}
J.~Wang, Q.~Zhu, Z.~Lin, Q.~Wu, Y.~Huang, X.~Cai, W.~Zhong, and Y.~Zhao, ``Sparse bayesian learning-based 3d radio environment map construction—sampling optimization, scenario-dependent dictionary construction and sparse recovery,'' \emph{IEEE transactions on cognitive communications and networking}, 2023.

\bibitem{RME_CS_2}
J.~Wang, Q.~Zhu, Z.~Lin, J.~Chen, G.~Ding, Q.~Wu, G.~Gu, and Q.~Gao, ``Sparse bayesian learning-based hierarchical construction for 3d radio environment maps incorporating channel shadowing,'' \emph{IEEE Transactions on Wireless Communications}, 2024.

\bibitem{CS}
J.~Zhang and B.~Ghanem, ``Ista-net: Interpretable optimization-inspired deep network for image compressive sensing,'' in \emph{Proceedings of the IEEE conference on computer vision and pattern recognition}, 2018, pp. 1828--1837.

\bibitem{CS_radio_review}
S.~K. Sharma, E.~Lagunas, S.~Chatzinotas, and B.~Ottersten, ``Application of compressive sensing in cognitive radio communications: A survey,'' \emph{IEEE communications surveys \& tutorials}, vol.~18, no.~3, pp. 1838--1860, 2016.

\bibitem{ZYBCS}
X.~Zhang, B.~Chen, W.~Zou, S.~Liu, Y.~Zhang, R.~Xiong, and J.~Zhang, ``Progressive content-aware coded hyperspectral snapshot compressive imaging,'' \emph{IEEE Transactions on Circuits and Systems for Video Technology}, 2024.

\bibitem{xie2022puert}
J.~Xie, J.~Zhang, Y.~Zhang, and X.~Ji, ``Puert: Probabilistic under-sampling and explicable reconstruction network for cs-mri,'' \emph{IEEE Journal of Selected Topics in Signal Processing}, vol.~16, no.~4, pp. 737--749, 2022.

\bibitem{alternating_optimization}
J.~C. Bezdek and R.~J. Hathaway, ``Convergence of alternating optimization,'' \emph{Neural, Parallel \& Scientific Computations}, vol.~11, no.~4, pp. 351--368, 2003.

\bibitem{dilated_conv}
F.~Yu and V.~Koltun, ``Multi-scale context aggregation by dilated convolutions,'' \emph{arXiv preprint arXiv:1511.07122}, 2015.

\bibitem{phy_RME}
S.~Zhang, B.~Choi, F.~Ouyang, and Z.~Ding, ``Physics-inspired machine learning for radiomap estimation: Integration of radio propagation models and artificial intelligence,'' \emph{IEEE communications magazine}, vol.~62, no.~8, pp. 155--161, 2024.

\bibitem{OCTUF}
J.~Song, C.~Mou, S.~Wang, S.~Ma, and J.~Zhang, ``Optimization-inspired cross-attention transformer for compressive sensing,'' in \emph{Proceedings of the IEEE/CVF conference on computer vision and pattern recognition}, 2023, pp. 6174--6184.

\bibitem{CPPNet}
Z.~Guo and H.~Gan, ``Cpp-net: Embracing multi-scale feature fusion into deep unfolding cp-ppa network for compressive sensing,'' in \emph{Proceedings of the IEEE/CVF Conference on Computer Vision and Pattern Recognition}, 2024, pp. 25\,086--25\,095.

\bibitem{CS_review}
J.~Zhang, B.~Chen, R.~Xiong, and Y.~Zhang, ``Physics-inspired compressive sensing: Beyond deep unrolling,'' \emph{IEEE Signal Processing Magazine}, vol.~40, no.~1, pp. 58--72, 2023.

\bibitem{FP_CS}
M.~Yako, Y.~Yamaoka, T.~Kiyohara, C.~Hosokawa, A.~Noda, K.~Tack, N.~Spooren, T.~Hirasawa, and A.~Ishikawa, ``Video-rate hyperspectral camera based on a cmos-compatible random array of fabry--p{\'e}rot filters,'' \emph{Nature Photonics}, vol.~17, no.~3, pp. 218--223, 2023.

\bibitem{ABG}
S.~Sun, T.~S. Rappaport, S.~Rangan, T.~A. Thomas, A.~Ghosh, I.~Z. Kovacs, I.~Rodriguez, O.~Koymen, A.~Partyka, and J.~Jarvelainen, ``Propagation path loss models for 5g urban micro-and macro-cellular scenarios,'' in \emph{2016 IEEE 83rd Vehicular Technology Conference (VTC Spring)}.\hskip 1em plus 0.5em minus 0.4em\relax IEEE, 2016, pp. 1--6.

\bibitem{molisch2012wireless}
A.~F. Molisch, \emph{Wireless communications}.\hskip 1em plus 0.5em minus 0.4em\relax John Wiley \& Sons, 2012, vol.~34.

\bibitem{MST}
Y.~Cai, J.~Lin, X.~Hu, H.~Wang, X.~Yuan, Y.~Zhang, R.~Timofte, and L.~Van~Gool, ``Mask-guided spectral-wise transformer for efficient hyperspectral image reconstruction,'' in \emph{Proceedings of the IEEE/CVF conference on computer vision and pattern recognition}, 2022, pp. 17\,502--17\,511.

\bibitem{ISTA}
D.~You, J.~Xie, and J.~Zhang, ``Ista-net++: Flexible deep unfolding network for compressive sensing,'' in \emph{2021 IEEE International Conference on Multimedia and Expo (ICME)}.\hskip 1em plus 0.5em minus 0.4em\relax IEEE, 2021, pp. 1--6.

\bibitem{CBAM}
S.~Woo, J.~Park, J.-Y. Lee, and I.~S. Kweon, ``Cbam: Convolutional block attention module,'' in \emph{Proceedings of the European conference on computer vision (ECCV)}, 2018, pp. 3--19.

\bibitem{dwconv}
F.~Chollet, ``Xception: Deep learning with depthwise separable convolutions,'' in \emph{Proceedings of the IEEE conference on computer vision and pattern recognition}, 2017, pp. 1251--1258.

\bibitem{CS_denoiser}
S.~H. Chan, X.~Wang, and O.~A. Elgendy, ``Plug-and-play admm for image restoration: Fixed-point convergence and applications,'' \emph{IEEE Transactions on Computational Imaging}, vol.~3, no.~1, pp. 84--98, 2016.

\bibitem{unet}
O.~Ronneberger, P.~Fischer, and T.~Brox, ``U-net: Convolutional networks for biomedical image segmentation,'' in \emph{Medical image computing and computer-assisted intervention--MICCAI 2015: 18th international conference, Munich, Germany, October 5-9, 2015, proceedings, part III 18}.\hskip 1em plus 0.5em minus 0.4em\relax Springer, 2015, pp. 234--241.

\bibitem{cao2024hybrid}
M.~Cao, L.~Wang, M.~Zhu, and X.~Yuan, ``Hybrid cnn-transformer architecture for efficient large-scale video snapshot compressive imaging,'' \emph{International Journal of Computer Vision}, vol. 132, no.~10, pp. 4521--4540, 2024.

\bibitem{DPM}
R.~Wahl, G.~W{\"o}lfle, P.~Wertz, P.~Wildbolz, and F.~Landstorfer, ``Dominant path prediction model for urban scenarios,'' in \emph{14th IST mobile and wireless communications summit}, 2005, pp. 1--5.

\bibitem{IRT}
T.~Rautiainen, G.~Wolfle, and R.~Hoppe, ``Verifying path loss and delay spread predictions of a 3d ray tracing propagation model in urban environment,'' in \emph{Proceedings IEEE 56th Vehicular Technology Conference}, vol.~4.\hskip 1em plus 0.5em minus 0.4em\relax IEEE, 2002, pp. 2470--2474.

\bibitem{SSIM}
M.~Shen, H.~Gan, C.~Ma, C.~Ning, H.~Li, and F.~Liu, ``Mtc-csnet: Marrying transformer and convolution for image compressed sensing,'' \emph{IEEE Transactions on Cybernetics}, 2024.

\bibitem{TL}
F.~Zhuang, Z.~Qi, K.~Duan, D.~Xi, Y.~Zhu, H.~Zhu, H.~Xiong, and Q.~He, ``A comprehensive survey on transfer learning,'' \emph{Proceedings of the IEEE}, vol. 109, no.~1, pp. 43--76, 2020.

\end{thebibliography}

\newpage

\end{document}